*Chapter *

# A novel approach to increase scalability while training machine learning algorithms using Bfloat – 16 in credit card fraud detection.

*By Bushra Yousuf , Rejwan Bin Sulaiman and Musarrat Saberin Nipun*

## Abstract

The use of credit cards has become quite common these days as digital banking has become the norm. With this increase, fraud in credit cards also has a huge problem and loss to the banks and customers alike. Normal software's not able to detect the fraudsters since they emerge with new techniques to commit fraud. This creates the need of using machine learning-based software to detect frauds. Currently, the machine learning software available focuses only on the accuracy of detecting the frauds but does not focus on cost or time factors. This thesis focuses on machine learning scalability for banks' credit card fraud detection systems. We have compared the existing machine learning algorithms and methods available with the newly proposed technique. The goal is to prove that using fewer bits for training a machine learning algorithm will result in a more scalable system i.e. that it will reduce the time and will also be less costly to implement.

# Table of Contents

# List of Figures

# List of Tables

# 1. Introduction

The use of banking sector plays a great role in everyone's life. We make use of banking products in the form of online usage and physical cards. Fraud occurs when someone steals your physical credit card or you share your credit card/bank details with deceitful people over the phone or internet.
When you share the credit card details with a corrupt person, this leads to online fraud, phishing and spamming. The chances of getting trapped for the customer and the banks by fraudsters are very high because of the large volume of the transactions occurring.[1]

The different types of frauds can include insurance fraud, credit card fraud, accounting fraud, which results in financial loss to the customers or banks. Thus, it's quite necessary to detect these kinds of frauds.

The detection of the frauds in accounting could be identified by looking at apparent and evident signals. High value transaction in unusual location or merchant needs additional verification. Traditional methods use rule-based systems to identity fraud practices, not focusing on dire situations, extreme imbalance of negative and positive instances. The rule-based fraud systems use methods which conducts various checks to identity different fraud frameworks, which is created manually by fraud analysts. [2]

The current traditional systems use around 400 different methods to validate a transaction. [2] The algorithms require adding more scenarios physically and can barely detect uncorrelated relationships. In addition to this, rule-based software frequently makes use of outdated software which can barely undertake the task of processing real-time data; required for the current market.

The detection of credit card fraud is complicated task, because of two reasons: (i) the behaviour of fraudsters is usually different every time (ii) the data is not balanced, i.e., the actual set of data sample outnumbers the limited data set samples meaning the real fraudulent cases.
When you provide an input of highly unbalances data to machine learning, the model becomes partial towards the actual dataset. As a consequence, it is more inclined to show a fraudulent record as authentic record. [3]

## 1.1 Overview of Machine Learning Algorithms

There are certain hidden and subtle events in the behaviour of a user, which is not obvious but can still indicate probable fraud. By using machine learning we can create algorithms that can process big data sets with different variables and help us identify the concealing correlations between the behaviour of the user and the fraudulent actions. One of the key strengths of machine learning over the conventional rule-based system is that is much faster in processing data and has less manual work involved.

For instance, machine learning algorithms incorporate well with behaviour analytics and this in return reduces the verification steps. Major financial institutions are already using machine learning technology to tackle the fraudsters. For example, MasterCard has combined AI and machine learning to process and track different variables like, time, transaction size, location, purchase data and device. The aforementioned system then evaluates the bank accounts performance through every set of operation and provide real-time reasoning i.e. if the transaction occurred is fraudulent or real. The objective of this project is to reduce the number of incorrect declines at the merchant payments. [4]

As per recent research, false declines made the loss of around $118 billion per year to the merchants and the client's loss is around $9 billion per year. [5] This is one of the key areas for the fraud in financial services. Therefore, preventing fraud is one of the most important factors for payment industries and banks. This shows that there is an immense need to develop an efficient machine learning based banks fraud detection system for credit card.

Machine learning needs large sets of data to train a model and training a model can take an extended time. A bank fraud detection system has real-time frauds appearing and this requires for a more efficient system. Also, a model can be so big that it cannot fit into the working memory of the training device.

Even if we decide to buy a big machine with huge memory and processing power, it is going to be somehow more expensive than using many smaller machines. In other words, vertical scaling is expensive. This creates a need for efficient and inexpensive method for machine learning scalability.

In this dissertation, the focus is on how to achieve machine learning scalability efficiently for banks fraud detection in credit card.

## 1.2 Problem Statement

The currently available machine learning algorithm for scalability focuses primarily on distributed computing such as using multicore processors, GPUs or HPCC which is quite costly. Moreover, banks credit card fraud detection system requires real-time data processing for anomaly cases on large data sets which creates the need for machine learning system.

A fraudster will not follow the same pattern repeatedly, it means we must train our system in a short period of time with a new dataset of cases. The current algorithms for machine learning require large period for training the data set and then testing the results and implementing it. In addition to this, the focus on algorithms of machine learning for bank fraud detection in credit cards is only on increasing the accuracy of algorithms not scalability.

By reducing the processing time i.e. making the system scalable, will indeed influence the accuracy. Regardless of that, the important goal is to train the model more quickly such that it can identify a portion of frauds in the meantime which would have been consumed in training the model. As research indicate, the undetected frauds can cost banks millions or billions of dollars in a limited amount of time.[5]

This implies the excessive time taken by training the models on large data sets for machine learning system for the fraud detection can cause huge loss to the bank for the time it is left being undetected.

## 1.3 Research Objective

The aim of the thesis is to present a solution for machine learning scalability in banks fraud detection system for credit cards which can process the data efficiently in less amount of time.

Currently, the machine learning algorithms comparison available is related to accuracy of different algorithms only.

- To research about the current machine learning scalability techniques available.
- To propose a new solution for machine learning scalability for banks fraud detection in credit cards which takes less amount of time in processing the data and supports efficiently to detect fraud in real time.
- To implement the algorithm and show test results in comparison with distributed computing scalability.

## 1.4 Report Overview

In recent years, fraud detection in credit card has increased tremendously, drawing the attention of most scholars and researchers. Researchers are trying to solve some methodological barriers that pose a limitation in ML real-time application. Various research has been done in different domains such as abnormal patterns detection [16], biometric identification [17,19], Diabetes Prediction [18], Anomaly detection [15], Pneumonia Detection [14], Predicting informational efficiency using deep Neural Network [13]. Despite these limitations, researchers are working to gain the ML power to detect frauds.

In this dissertation, one of the main goals is to study the existing algorithms of machine learning available for performance improvement in banks credit card fraud detection system i.e. logistic regression, decision trees, support vector machines and random forest algorithm. As research indicate, to further improve the performance of machine learning systems we require scalability, it is the need of the era. At present, the current studies are dedicated towards performance improvement for banks credit card fraud detection system by optimizing the algorithms only.

The key discussion is to propose a new training model for machine learning scalability which will improve the performance efficiency rather than focusing on algorithm optimization. It will focus on using the technique of low precision training method for machine learning scalability. Certain tests are conducted on the data to show the results which will indicate the performance improvement achieved through this method.

## 2. Literature Review

## 2.1 Introduction

The algorithms for machine learning, which are programs alter its mechanism to achieve better results as they are disclosed to additional information. In machine learning, the 'learning' part implies that such programs alter how they operate data over a period of time, in the same way how humans evolve information by learning.

Machine learning scalability means handling vast amount of data and performing a number of computations in a very cost-efficient and time saving manner. Below are the inherent advantages of giving importance to scaling.

- **Productiveness:** Currently, machine learning happens in the form of experiments, like finding a unique issue with a unique design (algorithm). A pipeline with quick executions of each stage (training, evaluation, and deployments) will enable us to try additional things and facilitate us to be more creative.
- **Portability:** It will be more beneficial if the training results and the model which is trained can be utilized by other teams for enhanced results.
- **Reduced Cost**: It is always a good option to do optimization for costs. If we scale, it will support us in utilizing the accessible resources to maximum. In addition, it will create a trade-off in accuracy and marginal cost.
- **Decrease Human Involvement**: The algorithms should be automated to the possible extent, so the humans can relax for some time and focus on other things.

For example, twenty fifth percent of engineers at Facebook assignment is to train models, training 550k models per month. Their on-line prediction service makes 5M predictions per second. Baidu Deep Search model training uses computing power of 200 TFLOP/s on a cluster of 128 GPUs. Thus, we can imagine how it is necessary for such corporations to scale with efficiency and why scalability in machine learning matters currently. [6]

The main objective of this literature review is to study the existing algorithms available for machine learning in banks credit card fraud detection system and explore the current possibility available for machine learning scalability.

Below is the overview for some most used machine learning algorithms for banks fraud detection.

## 2.2 Logistic Regression

It is a supervised classification methodology which returns the possibility of dependent binary variable that's speculated from the independent variable quantity of dataset. The logistic regression predicts the possibility of associate outcome that has two different values either 0 or 1, affirmative or no and true or false.

Logistic regression is like statistical regression however as in statistical regression a line is obtained, Logistic regression shows a curve. The utilization of 1 or many predictors or independent variables depends on what prediction is predicated, logistic regression produces curves that plots the values between 0 and 1.

Regression could be a regression model wherever the dependent variable is definite and examine the connection between various independent variables.[7]

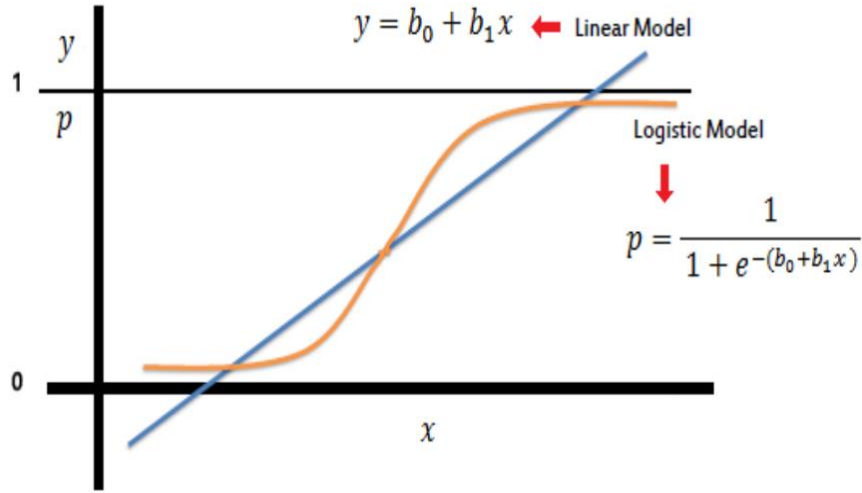

*Figure 1: Logistic Curve*

The above graph shows the difference between logistic regression and linear regression where logistic regression shows a curve but linear regression represents a straight line.

## 2.3 Decision Tree

It is an algorithm which approaches discontinuous-value target functions. It is denoted by a learned function. When doing inductive learning, such types of algorithms are well-known. They have been successfully used to a broad range of tasks.

We name label to a recent transaction stating whether it is real or fraudulent and for which the class label is not known. After wards the transaction value is tested with the decision tree, and then from root node to output for that particular transaction a path is detected. The decision rules identify the result of the information of the leaf node.

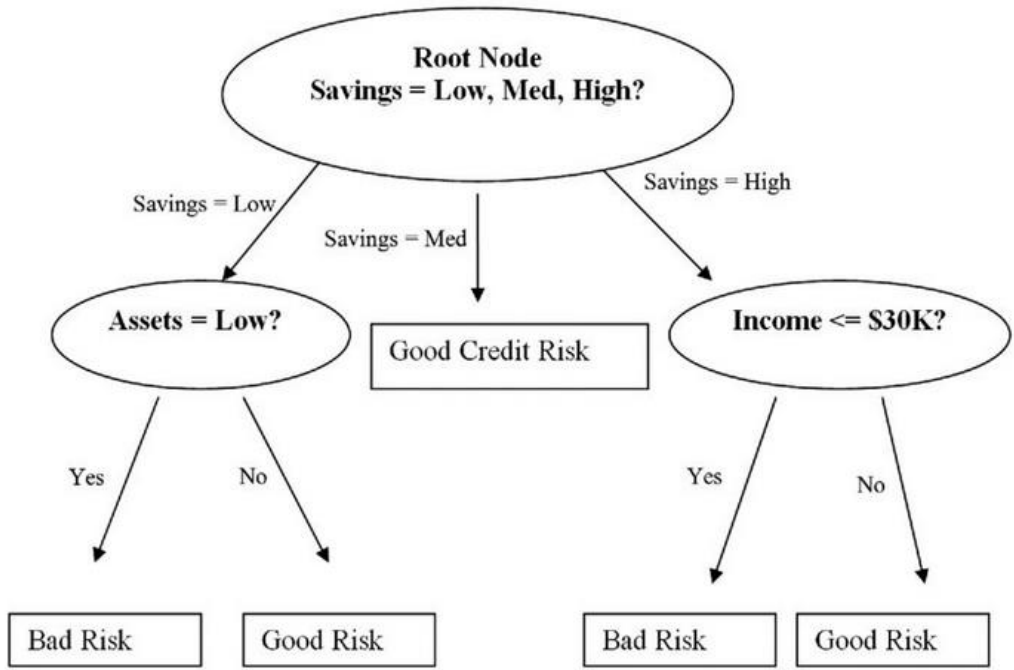

*Figure 2: Decision Tree*

Decision tree assists to calculate the best, expected and worse values for various situations, simplified to know and interpret and allows inclusion of new probable situations. Steps for creating a decision tree is that we evaluate the entropy of each attribute by utilizing the dataset in the problem. Afterwards the dataset is divided into different sets by utilizing the attribute who has maximum gain or minimum entropy to make a decision tree node consisting of those attributes. At the end, and lastly recursion is operated on subsets by again utilizing the remaining attributes to make a decision tree. [8]

| | Actual Condition | | Accuracy | |
|---|---|---|---|---|
| | Condition +ve | Condition – ve | 72% | |
| Predicted Condition +ve | 156 | 19 | Precision | False discovery rate |
| | | | 89% | 11% |
| Predicted Condition – ve | 51 | 24 | False omission rate | Negative predictive value |
| | | | 68% | 32% |
| Prevalence | Sensitivity, Recall TPR | Fallout FPR | +ve likelihood ratio | - ve likelihood ratio |
| 83% | 75% | 44% | 1.71 | 0.44 |
| | Miss Rate FNR | Specificity TNR | F1 Score | |
| | 25% | 56% | 71 | |

*Table 1: Confusion Matrix for Decision Tree*

In article (1), we found that the accuracy for decision tree is 72% and the likelihood ratio to detect false positive is 1.71% and for false negative is 0.44%. [8]

2.4 Support Vector Machines

It is also a supervised machine learning algorithm that is used for regression problems or classification. It makes use of a method which is known as 'kernel trick' that changes your data and later on the basis of these changes it locates an optimal boundary from the possible outputs.

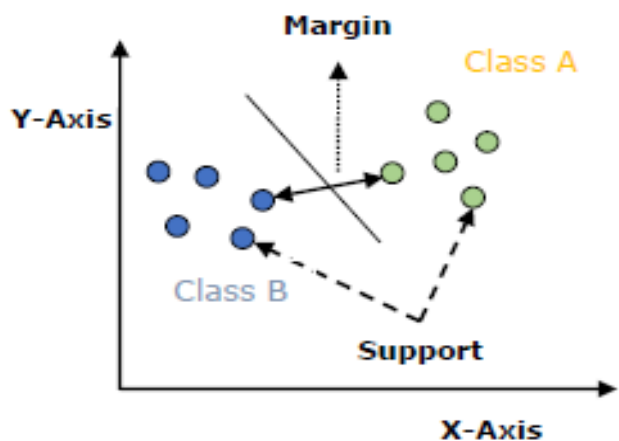

*Figure 3: SVM Model Graph*

In other words, it performs very intricate data transformation and then finds out to segregate the data based on the labels or the outputs which you have specified.

The model is a representation of multitude of classes in multidimensional space of a hyperplane. The hyperplane is then created in a repetitive way through SVM such that the error is reduced. The objective of this algorithm of SVM is to split the datasets into set of classes to locate a maximum marginal hyperplane (MMH). [9]

2.5 Random Forest

Random Forest is an algorithm for regression and classification. In a nutshell, it is a set of decision tree classifiers. It has an immense advantage over DT (decision trees) because it corrects the pattern of overfitting of their training set. Subset of the set of the training is sampled randomly in order to train every single tree so that a decision tree is created, every node then segregates a feature which is chosen from the random subset of the full set.

Training is extremely efficient even for larger data sets with different features and data because every tree is trained separately from the other trees.

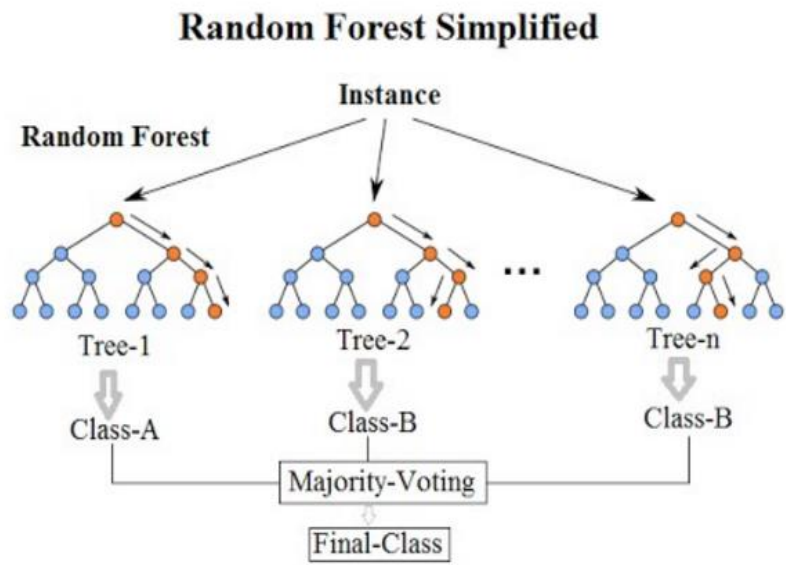

*Figure 4: Random Forest Classifier*

The Random Forest algorithm also ensures to give an appropriate estimate of the generalization issue and overfitting. The random forest ranks the importance of variables in classification or regression problem in a more natural way. [9]

In article (2), we observed that overall accuracy for Random Forest is 76% which is better than decision tree. The likelihood of false positive is 2.51% and false negative is 0.33% which is again less than decision trees. [8]

| | Actual Condition | | Accuracy | |
|---|---|---|---|---|
| | Condition +ve | Condition – ve | 76% | |
| Predicted Condition +ve | 163 | 12 | Precision | False discovery rate |
| | | | 93% | 7% |
| Predicted Condition - ve | 48 | 27 | False omission rate | Negative predictive value |
| | | | 64% | 36% |
| Prevalence | Sensitivity, Recall TPR | Fallout FPR | +ve likelihood ratio | - ve likelihood ratio |
| 84% | 77% | 31% | 2.51 | 0.33 |
| | Miss Rate FNR | Specificity TNR | F1 Score | |
| | 23% | 69% | 61 | |

*Table 2: Confusion Matrix for Random Forest*

Where

**Precision –** is the % of the true predictions

**FDR – is the** % of true predictions which are not true

**FOR – is the** % false predictions made which are true

**NPV – is the** % false predictions which are not true

**Miss Rate -** % of actual positive predicted wrongly?

2.6 Performance

Now that we have discussed how the machine learning algorithms work, we will discuss how they perform particularly for banks credit card fraud detection system.

| Model | Accuracy | Precision | Recall | TPR | FPR | F1-Score | G-Mean | Specificity |
|---|---|---|---|---|---|---|---|---|
| SC | 0.95270 | 0.95 | 0.95 | 0.9387 | 0.0335 | 0.95 | 0.9524 | 0.9664 |
| RF | 0.94594 | 0.95 | 0.95 | 0.9251 | 0.0335 | 0.95 | 0.9455 | 0.9664 |
| XGB Classifier | 0.94594 | 0.95 | 0.95 | 0.93197 | 0.0402 | 0.95 | 0.9457 | 0.9597 |
| KNN | 0.94256 | 0.91 | 0.91 | 0.9183 | 0.0335 | 0.91 | 0.942 | 0.9664 |
| LR | 0.93918 | 0.94 | 0.94 | 0.93877 | 0.0604 | 0.94 | 0.9391 | 0.9395 |
| GB | 0.93581 | 0.94 | 0.94 | 0.9183 | 0.0335 | 0.94 | 0.942 | 0.9664 |
| MLP Classifier | 0.93243 | 0.93 | 0.93 | 0.9387 | 0.0738 | 0.93 | 0.9323 | 0.9261 |
| SVM | 0.93243 | 0.93 | 0.93 | 0.9183 | 0.536 | 0.93 | 0.9321 | 0.9463 |
| Decision Tree | 0.90878 | 0.91 | 0.91 | 0.9047 | 0.0872 | 0.91 | 0.9086 | 0.9127 |
| Navies Bayes | 0.90540 | 0.91 | 0.91 | 0.85714 | 0.04697 | 0.91 | 0.9037 | 0.953 |

*Table 3: Performance evaluation of different classifiers*

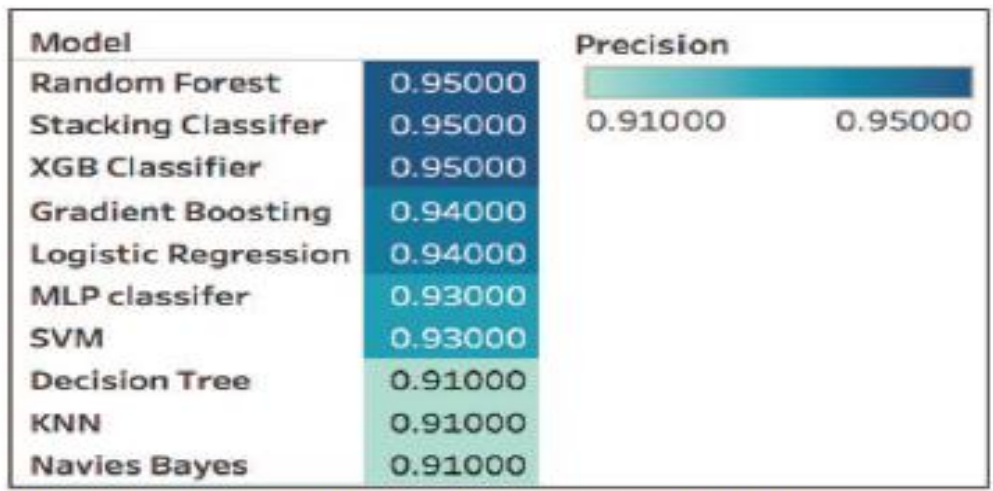

*Figure 5: Classifier ranking based on precision score*

We found that in article (3), the performance of Random Forest in 95%, followed by Logistic regression 94%, SVM is 93% and the lowest being Decision Tree 91%. [9]

In article (4), we found that the different works conducted in research papers with respect to machine learning algorithms used for credit card fraud detection system. The comparison chart below explains the frequency of the same technique being used, with the accuracy achieved along with the cost associated with it.

We found that supervised learning techniques are more frequently used than unsupervised techniques. As shown below, the accuracy of SVM, Bayesian Network, Fuzzy logic is high, but the cost is also high as well. Logistic Regression and Random Forest are the two algorithms with high accuracy and low cost as compare to other algorithms used for credit card fraud detection system. [10]

| Machine Learning algorithm | Type of technique: supervised (st) or unsupervised (ut) | Frequency | Accuracy | Coverage | Costs |
|---|---|---|---|---|---|
| Artificial Neural Network (ANN) | st | 40% | 2 | 2 | 3 |
| Decision Tree (DT) | st | 38% | 2 | 2 | 3 |
| Support Vector Machine (SVM) | st | 34% | 3 | 3 | 3 |
| Genetic algorithm (GA) | ut | 26% | 2 | 2 | 1 |
| K-nearest Neighbor (KNN) | st | 20% | 2 | 2 | 3 |
| Bayesian Network | st | 16% | 3 | 2 | 3 |
| Hidden Markov Model (HMM) | ut | 16% | 1 | 1 | 3 |
| Logistic regression | st | 16% | 3 | 2 | 2 |
| Random forest | st | 16% | 3 | 2 | 2 |
| Naïve Bayes | st | 14% | 2 | 2 | 3 |
| Self-organizing map (SOM) | ut | 8% | 2 | 1 | 3 |
| Fuzzy Logic Based system (FL) | st | 8% | 3 | 2 | 3 |
| Artificial Immune System (AIS) | st | 8% | 2 | 3 | 1 |
| DBSCAN | ut | 4% | 3 | 2 | 3 |

*Table 4: Classification of algorithms based on frequency and accuracy*

The results discussed only focuses on the accuracy of algorithm without taking in consideration how much time was consumed in training the data set to detect the fraud in credit cards.

At present, the main focus on achieving machine learning scalability is primarily on distributed computing for machine learning scalability.

## 2.7 Distributed Computing for Machine Learning

By using parallel processing, we can process vast amount of data because it allows for accelerated execution of tasks and utilizes less amount of computer resources. We can do parallel processing in the form of horizontal or vertical scaling.

While using vertical parallel processing, we increase the storage and power of processing of one machine, but in horizontal parallel processing we distribute the tasks on different machines. We can achieve vertical parallel processing by using HPCC and GPU.

The two types of distributed learning algorithms in terms of fragmentation are stated below:

- The horizontal fragmentation is when the subsets of different instances is located on various sites.
- The vertical fragmentation is when the subsets of instances of various attributes are located at various sites.

Some examples of distributed machine learning algorithms are in advertising or healthcare where an uncomplicated application can have multitudes of data. The data is so vast, that programmers has to routinely re-train the data set to ensure that workflow is not interrupted, and the load is distributed in parallel.

For instance, the MapReduce was created to permit flow of automatic parallelization and distribution of vast scale special-purpose computations which can process huge amounts of raw data, like searched documents or request web logs and calculate different types of derived data.

In article (5), we found that one study divided the large data set of credit card transaction into smaller sets and then applied mining techniques to generate more classifiers in parallel. They also combined both results by using meta learning and by identifying their behaviour they generated a metaclassifier. They executed the metaclassifiers in parallel to then achieve scalability. [11]

## 2.8 Applications of Distributed Computing for Machine Learning

Microsoft also have Distributed ML Toolkit, that consists of both system innovations and algorithmic. Microsoft framework of Distributed Machine Learning Toolkit maintains an interface for parallelisation of data, heterogenous data structure for large storage, scheduling for large training model, and automated pipelining for more efficient training. [12]

Machine learning scalability, which is focusing primarily on distributed computing, is not cost-effective. In addition, it is very complex to write and run a distributed ML algorithm and to develop distributed ML packages becomes tough due to of dependency of platform. Since, there are no conventional measures to assess distributed algorithms. Many machine learning researchers presume that the current measures specified against classical ML methods exhibit little reliability.

Currently, there has been some work done online on datasets provided free on websites like Kaggle to detect frauds in credit cards. There is a dataset which is available online with customer records of showing fraud occurred on their credit cards. We will utilize the same dataset for the thesis as well in order to provide a coherent comparison.

## 2.9 Conclusion

Machine learning scalability is the demand of the era for banks credit card fraud detection system. As detecting frauds on large data sets require tremendous amount of resources and time, which is both of essence.

The goal of this literature review was to compare the currently used algorithms for machine learning in banks fraud detection system and the ongoing machine learning scalability method available. The literature review implies that there is a crucial need to further improve the machine learning scalability as relying only on parallelism is not a feasible option due to cost constraints.

## 3. Research Methodology

### 3.1 Introduction

The objective of this chapter is to discuss the description of the research process. It will provide information for choosing the research approach and research method for this dissertation. It will also discuss in detail the analysis method and tool used. The focus is to present a solution approach for machine learning scalability in banks credit card fraud detection system. To conclude this chapter, it will also enclose the assumptions, constraints and limitations concerning this paper.

### 3.2 Research Approach

The research approach utilized in this paper is quantitative empirical research, since the paper will present the results in the form of numerical data. The quantitative empirical research approach applied here will help us create a statistical analysis in comparing the results of existing algorithms available for machine learning scalability in banks credit card fraud detection system with the solution which we are proposing.

### 3.2 Justification for Research Approach

The reason to use quantitative empirical research is because it can be converted into usable statistics. It has a more structured approach and the statistics derived from this approach will help us determine the probability of the success of our solution presented. It will also help us track changes over the time in the project due to statistical data, if any. It will support in testing the hypothesis, where we believe that the proposed solution will increase the machine learning scalability in credit cards fraud detection system at the expense of reducing some accuracy.

### 3.3 Research Method

The research method used is experiments conducted on the machine learning algorithms and compare the results of previous results with the proposed solution in this paper. As we are doing comparison, statistical

13 | P a g e

inference is used to determine how one method is better than the other. The experiments conducted will to identify the relationship between different variables used for the system. We are using the quasi-experimental method, which means we will not utilize random assignment of variables for the system i.e. it will be fixed.

### 3.4 Data Collection

The data for this system is collected from existing database of credit cards of the banks, where it will consist of true positive data where the fraudulent activity was detected and also such cases where there was no fraud detected in order to be able to train the machine learning model for false positive as well. The data is stored in the form of CSV file.

### 3.5 Data Analysis

The analysis of the data is done by using schema-based validation in TensorFlow software. Then results are presented in the form of graphs and charts to show the comparison between the both methods used for machine learning scalability in banks credit card fraud detection system.

### 3.6 Assumptions

- The data which is used for training the machine learning model is supervised data.
- We assume that the data used for this system is enormous and it is enough to train the machine learning model correctly for credit card fraud detection.
- The dataset which is used for machine learning is imbalanced.
- The data which is available will be compatible with all the tools which will be used for the system.

### 3.7 Constraints

- As training large data sets require huge processing power i.e. GPUs, Supercomputers or HPCC (High-Performance Computing Cluster). Currently it is not possible to use such computational power due to cost and time constraints.
- Since the data set required for training such model is huge, it will not be possible to arrange storage for keeping such large data set.
- The data provided for machine learning should have actual fraud occurring and also where no fraud has been taken place to be able to properly train the model.

### 3.8 Limitations

- It is not possible to gather millions of records of data to be able to train the machine learning model accurately.
- The results provided are limited as they are presented in numerical terms only.
- We have limited processing power available so training the model can take more time.
- There is a potential for biased results as the results are dependent on the trained model data.

## 3.9 Tool

The tool used are TensorFlow which is a software library for high-speed numerical calculation. The architecture provides flexibility and uncomplicated deployment of computation across different platforms (CPUs, GPUs, TPUs), and from PC to set of servers to mobile and other edge devices.

In addition to this, Python and R is used as a programming language, python is quite popular in machine learning and data science whereas R is commonly used for statistical computing and data analysis. We will also be using the Google Cloud Platform (GCP) which is an easy platform to use and will provide multiple functionalities under one umbrella.

## 3.10 Proposed Method

The idea is to implement low precision training method. Generally, machine learning frameworks, by default, use 32-bit floating point precision for inference and training the models. There is evidence that we can use to lower numerical precision (like 16-bit for training, and 8-bit for inference) at the cost of minimal accuracy.

We are using float16 bit which will use 16-bits to represent the floating-point number. It is also known as half-precision float. The first is used for sign bit, the next 5-bits are used for exponent and the last 10-bits for fraction/mantissa. It is easily available to use in numpy library of python.

Reducing the precision will right away lead to reduced memory footprint, better bandwidth utilization, improved caching, throughput will be doubled, and sped up model performance (hardware can perform more operations per second on low precision operands).

# 4. System Design

This chapter will focus on the main design and implementation of the machine learning system for credit card fraud detection. We will also show the blueprint of the design by creating a prototype of the system.

## 4.1 Prototype

The first step is to gather raw data which can be then trained by the machine learning model, as shown in figure 6 below. The process diagram is discussed in detail:

1. **Stage One:**
    i. **Data Wrangling:**
        - We have to identify and source the data from the datasets and also reduce the set to only required data.
        - In addition to this, parsing has to be applied so the data can have a structure and can also be stored for future use.
        - We have to manipulate the raw data in such a way that it can be used to train the machine learning model.

    ii. **Data Cleaning**
        - We have to ensure that the different data sets collected are all in the same format, i.e.: data collected from images, audios and texts has to be put in the same format.
        - At this step, the data must be checked thoroughly for any typing mistakes, removing any unnecessary variables, remove any rows/columns which has many missing or empty values.
        - The data has to be verified such that it makes sense.
        - We have to select the important features from the set of different features which we require for prediction and this will also help in efficient computations and low-memory usage.

    iii. **Pre-processing:**
        - At this step, we have to normalize the data.
        - The data is transformed and evenly distributed so it can be used for machine learning model.

2. **Stage Two:**
    i. **Low-Precision Method:**
        - We will apply the proposed method here on the data set by using float16, it will be implemented on the data set in the TensorFlow with the help of libraries.
        - This will convert the 32-bit or 62-bit data into 16-bit data on which then the machine learning model is trained.

3. **Stage Three:**
    i. **Select Learning Technique:**
        - We will select supervised learning technique as it is easy to implement, because the data can be labelled and made into different classes. There will be only two classes 0 and 1. We will train our data set to identify the fraudulent cases from non-fraudulent cases.

- Since the data provided is not balanced, meaning that the number of fraudulent cases will not be the same as non-fraudulent cases, before we train our model, we have to balance the data.

ii. **Select Sampling Method:**
- There are multiple sampling methods available, but we will choose SMOTE (Synthetic minority oversampling technique), where we will synthetically make the fraudulent and non-fraudulent transactions equal in our data set.

iii. **Select Classification Technique:**
- We will choose the random forest classifier which is advanced version of decision trees, the reason for choosing this is established in the literature review as it is most accurate algorithm in comparison with other algorithms for machine learning.

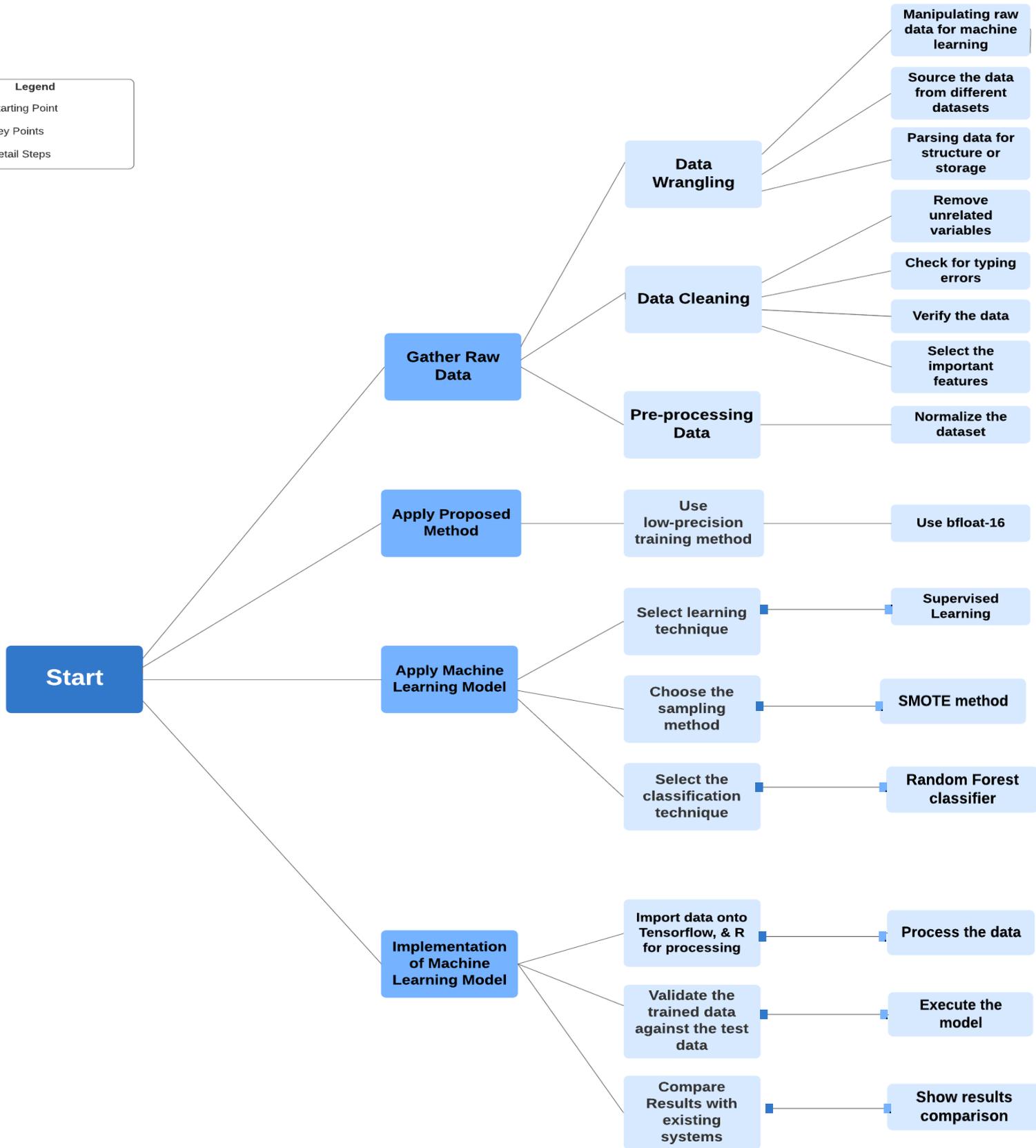

*Figure 6: Detailed Process for Machine learning scalability for credit card fraud detection system*

4. **Stage Four:**

    i. **Process the Data:**
        - We will load our data into TensorFlow and make it ready for processing.
        - R will also be used for statistical analysis.

    ii. **Run Validation Tests:**
        - We will execute our model and run validation tests of the trained model against the test model to identify the confusion matrix.

    iii. **Compare Results:**
        - We will show results of random forest algorithm implemented on the same data and show comparison in terms of time taken by training the same model with 32-bits and the resources utilized along with the confusion matrix.
        - This will help us understand how machine learning scalability was achieved.

## 4.2 Model Design

This part will cover the different model design components in detail. As the thesis focuses on machine learning scalability for credit card fraud detection, the most important factor is the database on which the training model will be applied. This module will focus on database design, interface design i.e. what is the environment in which the algorithm will be applied by using different programming languages.

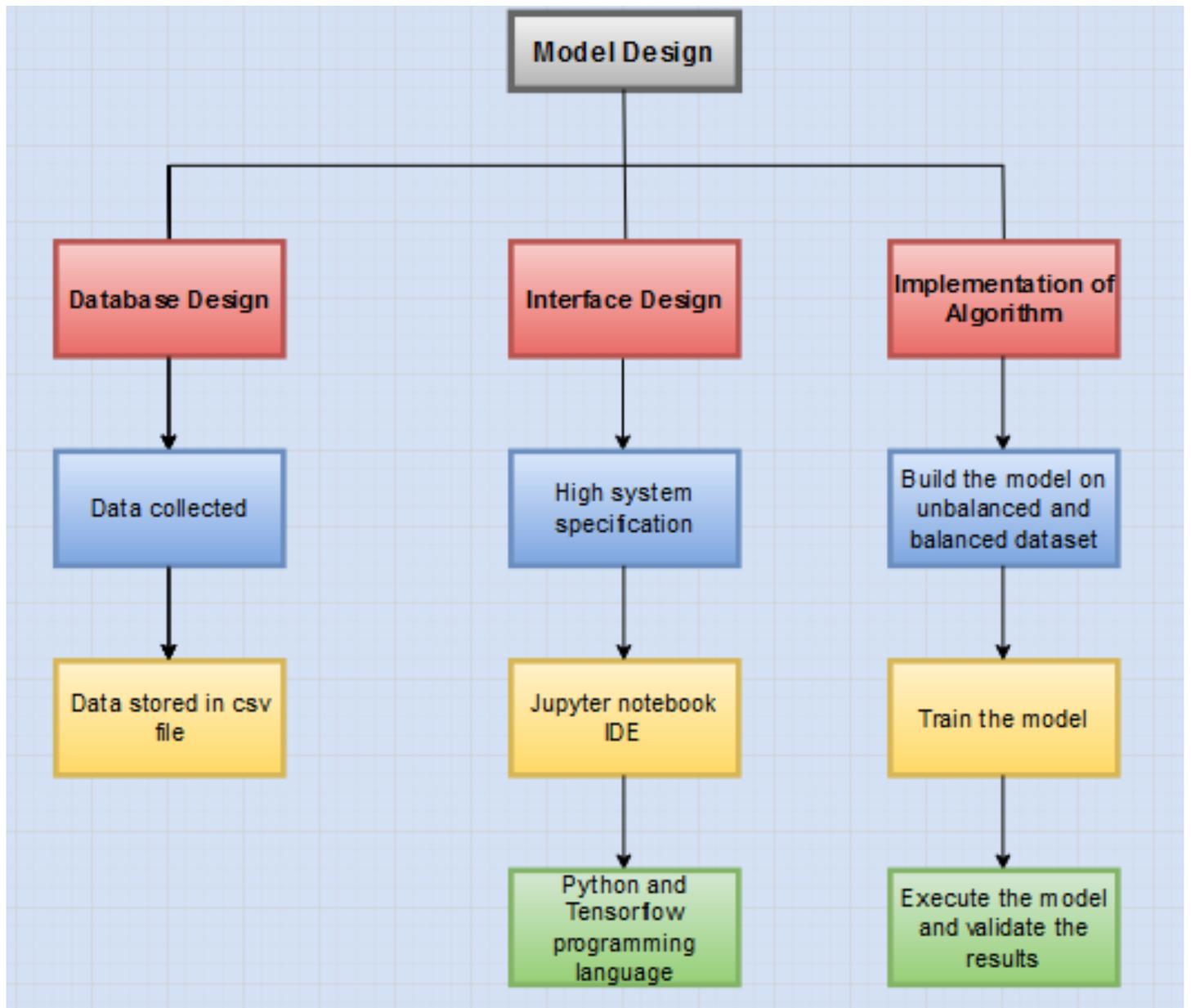

*Figure 7: Model design flow process for machine learning scalability for credit card fraud detection*

As depicted in figure 7, the first step is creating the database design which has to be stored in a format which can be manipulated by the programming. We will also discuss each part in more detail.

*4.2.1 Database Design*

The dataset contains transactions occurred by European credit card holders in September 2013. This dataset presents transactions that transpired over two days, where we have 492 frauds out of 284,807 transactions.

1. Data Analysis

To examine the data, we will show how the data is represented and enumerate the computation for different factors. The dataset is highly unbalanced, the positive class (frauds) account for 0.172% of all transactions. It consists only of numerical input variables which is an outcome of using a PCA (Principal component analysis) transformation.

Due to confidentiality issues and ensure privacy, original features and more background information about the data are not provided. Features V1, V2, ... V28 are the principal components obtained with PCA, the only features which have not been transformed with PCA are 'Time' and 'Amount'. Feature 'Time' contains the seconds elapsed between each transaction and the first transaction in the dataset. The feature 'Amount' is the transaction Amount, this feature can be used for example-dependant cost-sensitive learning. Feature 'Class' is the response variable, and it takes value 1 in case of fraud and 0 otherwise

|   | Time | V1 | V2 | V3 | V4 | V5 | V6 | V7 | V8 | V9 | ... | V21 | V22 | V23 | V24 |
|---|------|----|----|----|----|----|----|----|----|----|----|-----|-----|-----|-----|
| 0 | 0.0 | -1.359807 | -0.072781 | 2.536347 | 1.378155 | -0.338321 | 0.462388 | 0.239599 | 0.098698 | 0.363787 | ... | -0.018307 | 0.277838 | -0.110474 | 0.066928 | 0.12 |
| 1 | 0.0 | 1.191857 | 0.266151 | 0.166480 | 0.448154 | 0.060018 | -0.082361 | -0.078803 | 0.085102 | -0.255425 | ... | -0.225775 | -0.638672 | 0.101288 | -0.339846 | 0.16 |
| 2 | 1.0 | -1.358354 | -1.340163 | 1.773209 | 0.379780 | -0.503198 | 1.800499 | 0.791461 | 0.247676 | -1.514654 | ... | 0.247998 | 0.771679 | 0.909412 | -0.689281 | -0.32 |
| 3 | 1.0 | -0.966272 | -0.185226 | 1.792993 | -0.863291 | -0.010309 | 1.247203 | 0.237609 | 0.377436 | -1.387024 | ... | -0.108300 | 0.005274 | -0.190321 | -1.175575 | 0.64 |
| 4 | 2.0 | -1.158233 | 0.877737 | 1.548718 | 0.403034 | -0.407193 | 0.095921 | 0.592941 | -0.270533 | 0.817739 | ... | -0.009431 | 0.798278 | -0.137458 | 0.141267 | -0.20 |

5 rows × 31 columns

*Table 5: Representation of the dataset for credit card frauds*

As depicted in table 5, the header is only showing the first 5 entries of the dataset to show how data is stored using the PCA transformation which hides the original information due to confidentiality and privacy issues, to protect the identity of the customers.

|        | Time          | Amount        | Class         |
|--------|---------------|---------------|---------------|
| count  | 284807.000000 | 284807.000000 | 284807.000000 |
| mean   | 94813.859575  | 88.349619     | 0.001727      |
| std    | 47488.145955  | 250.120109    | 0.041527      |
| min    | 0.000000      | 0.000000      | 0.000000      |
| 25%    | 54201.500000  | 5.600000      | 0.000000      |
| 50%    | 84692.000000  | 22.000000     | 0.000000      |
| 75%    | 139320.500000 | 77.165000     | 0.000000      |
| max    | 172792.000000 | 25691.160000  | 1.000000      |

*Table 6: Statistical insight into the dataset*

As shown above, only 3 features of the dataset have not gone under PCA transformation as they are the critical information on which the machine learning model is built and trained. The maximum amount which was stolen due to fraudulent activity is 25691 USD. The fraudulent class is identified by using 0 and 1, if there was no fraud the value is 0 and if there was fraud the class value is 1.

Synopsis:
- The transaction amount is comparatively small. The mean of all the mounts made is approximately USD 88.
- After checking, we came to know that there are no "Null" values, which means we don't have to replace or add any values.
- Most of the transactions were non-Fraud (99.83%) of the time out of 280,000 while Fraud transactions occurs (0.17%) of the time in the data frame. This shows that the data has a large disparity in non-fraud to fraud ratio.

1. Data Coherency

We must ensure that the dataset does not consist of any null values, which can affect computations and show incorrect values. The dataset is stored in a csv file format. The below figure shows that there are no null values in the dataset which ensures data coherency.

```
 6    V6      284807 non-null  float64
 7    V7      284807 non-null  float64
 8    V8      284807 non-null  float64
 9    V9      284807 non-null  float64
10    V10     284807 non-null  float64
11    V11     284807 non-null  float64
12    V12     284807 non-null  float64
13    V13     284807 non-null  float64
14    V14     284807 non-null  float64
15    V15     284807 non-null  float64
16    V16     284807 non-null  float64
17    V17     284807 non-null  float64
18    V18     284807 non-null  float64
19    V19     284807 non-null  float64
20    V20     284807 non-null  float64
21    V21     284807 non-null  float64
22    V22     284807 non-null  float64
23    V23     284807 non-null  float64
24    V24     284807 non-null  float64
25    V25     284807 non-null  float64
26    V26     284807 non-null  float64
27    V27     284807 non-null  float64
28    V28     284807 non-null  float64
29    Amount  284807 non-null  float64
30    Class   284807 non-null  int64
dtypes: float64(30), int64(1)
memory usage: 67.4 MB
```

*Table 7: Non-null values of dataset*

### 4.2.2 Interface Design

In this part, we will discuss about system specification on which the software will run. The environment on which the programming language is used.

1. System Specification:

    The interface is run on 64-bit Windows 10 with intel core i3, as shown in below figure with specification. The system is currently run on CPU and not on GPU due to resource and cost limitation.

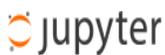

| Windows edition | |
|---|---|
| Windows 10 Pro | |
| © 2020 Microsoft Corporation. All rights reserved. | 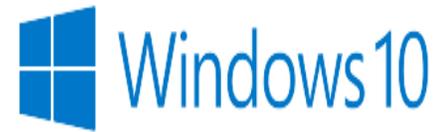 |

| System | |
|---|---|
| Processor: | Intel(R) Core(TM) i3-6006U CPU @ 2.00GHz  2.00 GHz |
| Installed memory (RAM): | 4.00 GB |
| System type: | 64-bit Operating System, x64-based processor |
| Pen and Touch: | No Pen or Touch Input is available for this Display |

*Figure 8: System specification with processor speed*

2. Jupyter Notebook (IDE)

   The Jupyter Notebook is an open-source web application that allows you to create and share documents that contain live code, equations, visualizations, and narrative text. Uses include data cleaning and transformation, numerical simulation, statistical modelling, data visualization and machine learning. We are using python and TensorFlow as part of Jupyter Notebook IDE.

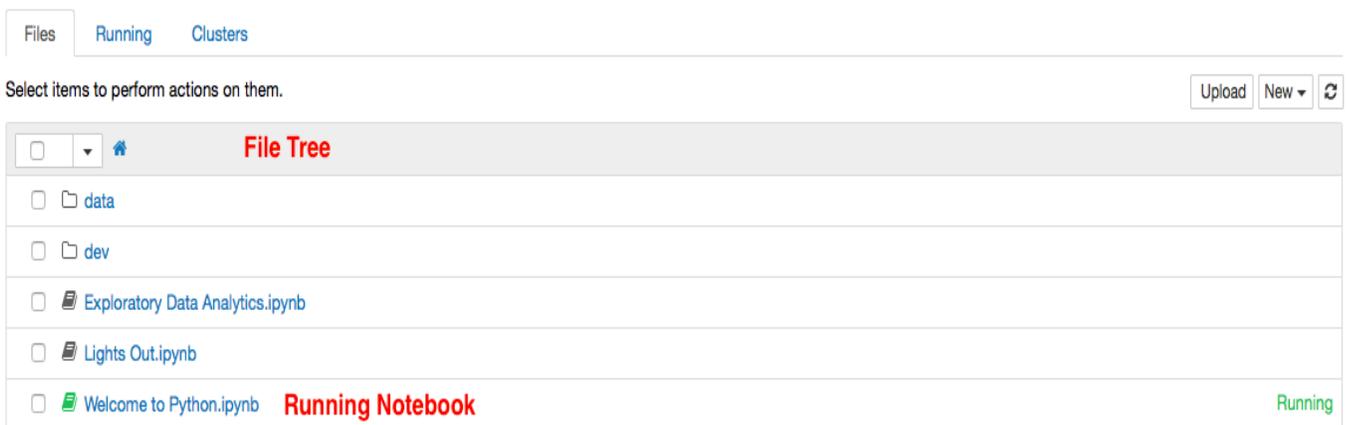

*Figure 9: Visualization of Jupyter Notebook*

Python is an interpreted, high-level, and general-purpose programming language used for multiple purpose. We are using python for machine learning. TensorFlow is an open-source library for numerical computation and large-scale machine learning. The proposed solution works only in TensorFlow and therefore we have to incorporate the use of TensorFlow along with python.

*4.2.3 Implementation of the Model*

This section will particularly cover the details of the implementation of applying bfloat – 16 techniques for the machine learning scalability for credit card fraud detection. We will also discuss the pre-processing steps required to implement before we apply the proposed solution.

1. Data Visualization

We are using data visualization techniques available in python to see the other features of the dataset. Below graphs show the distribution chart for each colour including Time and Amount.

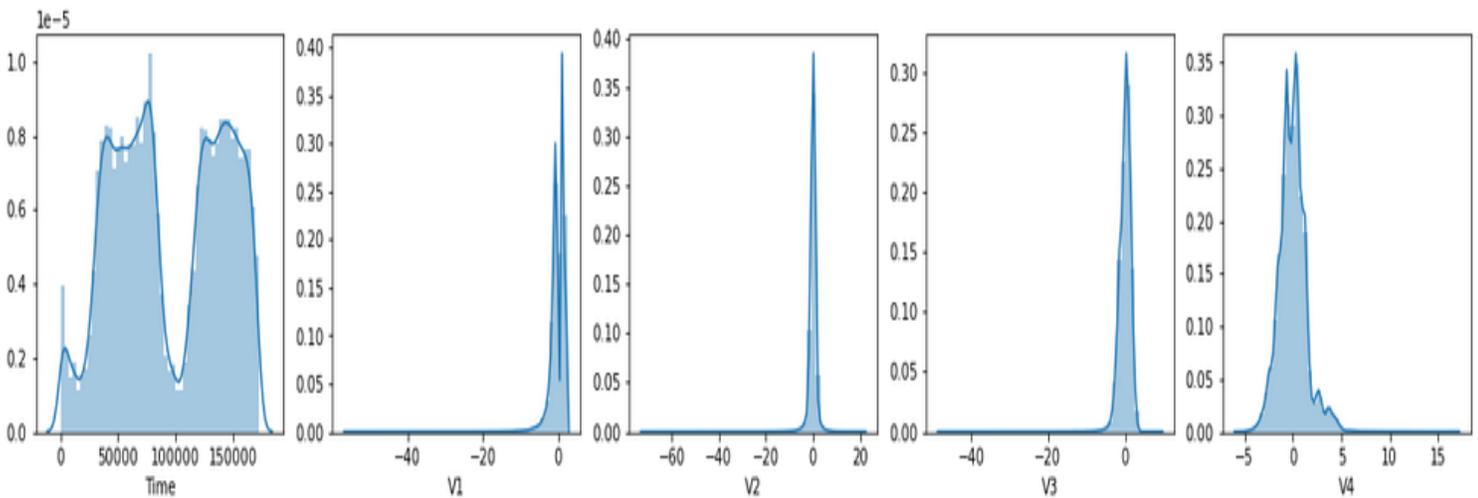

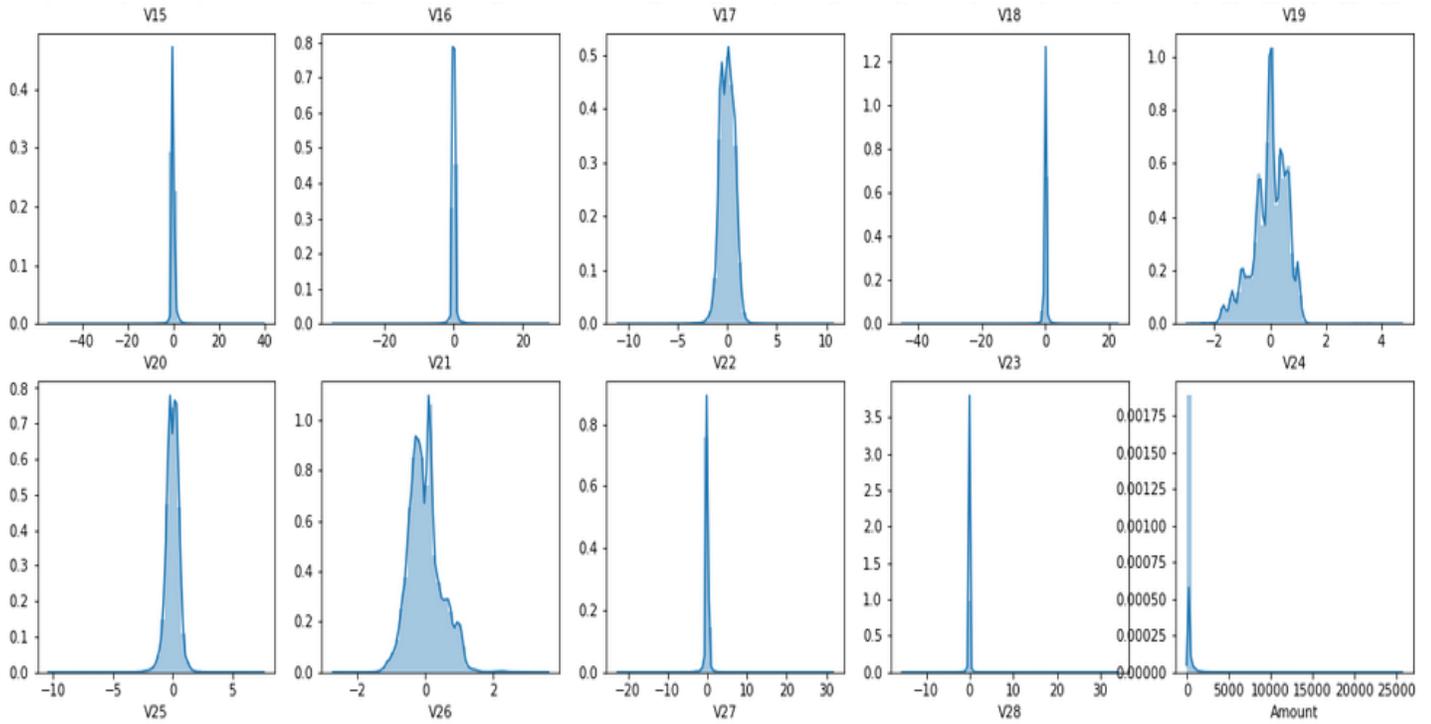

*Figure 10: Graphs depicting the columns of dataset*

As other columns information has been hidden due to the PCA transformation, we can focus visualizing Time and Amount more specifically.

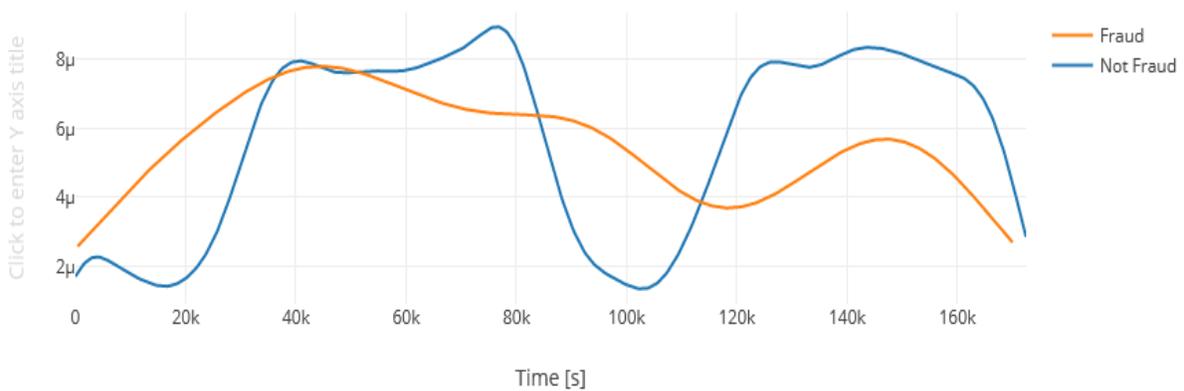

*Figure 11: Credit card transactions graph showing the fraud and non-fraudulent transactions*

We can deduce that, fraudulent transactions have more even distribution more even than non-fraudulent transactions. Fraudulent transactions are equally distributed in time.

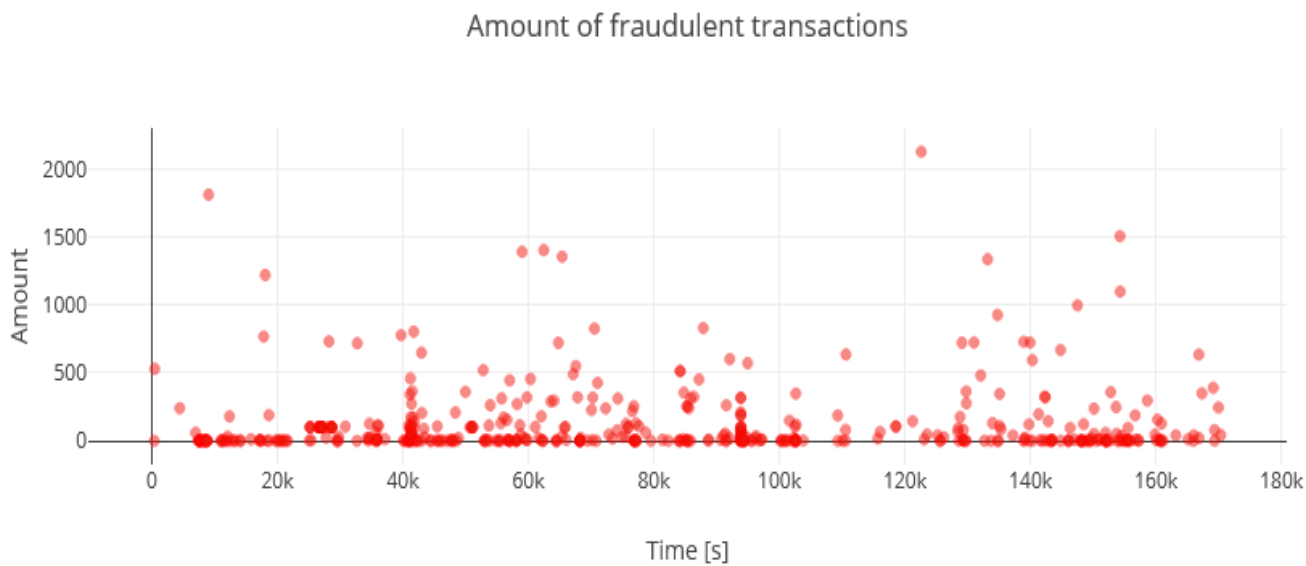

*Figure 12: Fraudulent transactions over time and amount*

Fraudulent transactions (amount) against time. The time is shown is seconds from the start of the time (totally 48h, over 2 days). This shows that many frauds have occur during the night-time rather than the daytime.

2. Supervised Machine Learning – Unbalanced Data

    We will implement supervised machine learning technique, as the dataset we have is divided into different classes and values are known. A model that is trained on a set of properly "labelled" transactions. Each transaction is marked as either fraud or non-fraud. Supervised machine learning model accuracy is directly correlated with the amount of clean, relevant training data.

    We can observe from the below distribution graph that data is highly unbalanced. We have almost 99% non-fraudulent cases represented by class 0, marked in colour blue and remaining 1% as fraudulent cases represented by class 1, marked in colour red.

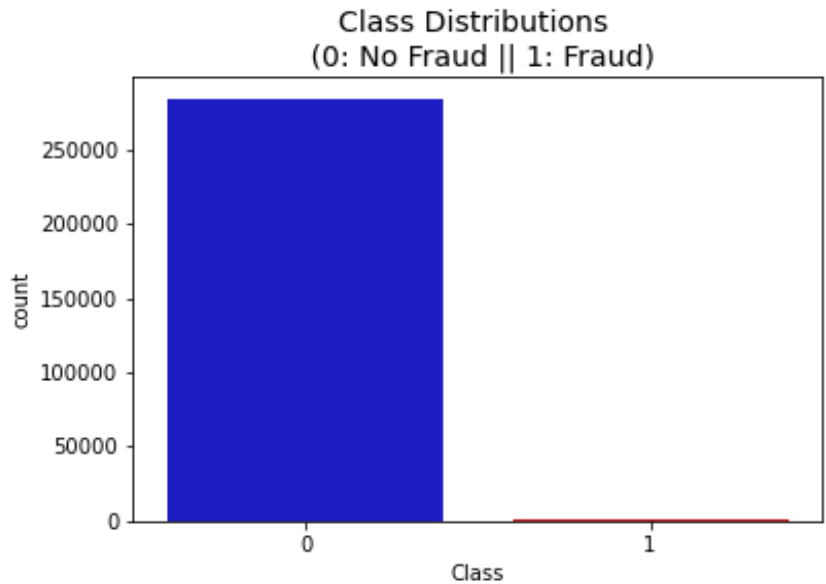

*Figure 13: Distribution of fraud and non-fraud class*

- We will divide the dataset into predictors and target data for the same unbalanced data and train the model.

```
y = df_credit.iloc[:,30]
X = df_credit.iloc[:,0:30]

print(X.columns)
print(y.head(3))

Index(['Time', 'V1', 'V2', 'V3', 'V4', 'V5', 'V6', 'V7', 'V8', 'V9', 'V10',
       'V11', 'V12', 'V13', 'V14', 'V15', 'V16', 'V17', 'V18', 'V19', 'V20',
       'V21', 'V22', 'V23', 'V24', 'V25', 'V26', 'V27', 'V28', 'Amount'],
      dtype='object')
0    0
1    0
2    0
Name: Class, dtype: int64
```

*Figure 14: Screenshot from python for dataset column headers*

- We will use the classification algorithm Random Forest to generate the confusion matrix for the unbalanced data.

    As shown below, we have used python to calculate the accuracy and generate the confusion matrix.

```
RF - Accuracy -  0.9995201479348805
RF:
 [[85286     9]
 [   32   116]]
RF - AUC:  0.9444259034226207
Random Forest:
 Precision:  [0.99962493 0.928     ] Recall:  [0.99989448 0.78378378]
```

*Figure 15: Screenshot from python interface*

| Confusion Matrix – Unbalanced Data ||
|:---:|:---:|
| 85286 | 9 |
| 32 | 116 |
| **Not Fraud** | **Fraud** |

(Rows: Actual — Not Fraud, Fraud; Columns: Predicted)

*Table 8: Confusion matrix for unbalanced data*

**Random Forest Classifier**

| | |
|---|---|
| **Accuracy** | 0.999520148 |
| **Precision** | 0.99962493 |
| **Recall** | 0.99989448 |

*Table 9: Randorm Forest classifier for unbalanced dataset*

As we can identify from the confusion matrix, the true negative value is quite high. This is because we have an unbalanced dataset.

3. Supervised Machine Learning – Balanced Data

There are two types of resampling methods to deal with imbalanced data, one is under sampling and another one is over sampling.

- Under sampling: We take random values from non-fraud observations to match the amount of fraud observations. We are randomly inserting a lot of data and information. It is also called Random Under Sampling.
- Over sampling: We take random values from fraud cases and copy these observations to increase the amount of fraud samples in our data.
- Synthetic Minority Oversampling Technique (SMOTE): We adjust the data imbalance by oversampling the minority observations (fraud cases) using nearest neighbors of fraud cases to create new synthetic fraud cases instead of just coping the minority samples.

We are using SMOTE technique to balance our data and then identify the confusion matrix and accuracy. The SMOTE technique is the best option as it creates new synthetic values instead of normal duplication.

```python
# Oversampling and balancing using SMOTE
#creating equal number of frauds and non-frauds

sm = SMOTE(random_state=42)
X_bal, y_bal = sm.fit_sample(X_train, y_train)

columns = X_train.columns
X_bal = pd.DataFrame(data = X_bal, columns = columns)

print(X_train.shape)
print(X_bal.shape)
print(np.unique(y_bal, return_counts=True))

#We now have equal occurences of Fraud and Non Fraud cases in y_bal
```
```
(199364, 30)
(398040, 30)
(array([0, 1], dtype=int64), array([199020, 199020], dtype=int64))
```

*Figure 16: Running of SMOTE on python*

- We will use the classification algorithm Random Forest to generate the confusion matrix for the balanced data.

```
RF - Accuracy -  0.9996137776061234
RF:
 [[85293     2]
 [   31   117]]
RF - AUC:  0.9546151433102603
Random Forest:
 Precision:  [0.99963668 0.98319328] Recall:  [0.99997655 0.79054054]
3.671875
```

*Figure 17: Screenshot from python interface*

| | Confusion Matrix | |
|---|---|---|
| No | 85293 | 2 |

|  | Not Fraud | Fraud |
|---|---|---|
| **Fraud** | 31 | 117 |

*Table 10: Confusion matrix for unbalanced data*

| Random Forest Classifier (SMOTE) ||
|---|---|
| Accuracy | 0.9996137776 |
| Precision | 0.99963668 |
| Recall | 0.99997655 |

*Table 11: Calculation of Accuracy & Other parameters*

We have higher accuracy and precision as compare to the unbalanced dataset.

## 5. Results & Comparison

In this chapter we will share the results of the algorithm applied and compare the results with previous outcome. It will present a clear understanding on how implementing float16 data type can provide scalability.

### 5.1 Descriptive Statistics

We will replicate the steps mentioned in chapter 4 again, but before we do that, we will use float16 bits data type on the data file which was stored in csv format. We have used Google Cloud Platform to run the algorithm with same machine specification for better storage and future work.
The data file was stored in Google cloud buckets storage and accessed from Jupyter Notebook using python programming language.

|  | Confusion Matrix | |
|---|---|---|
| **Not Fraud** | 85287 | 8 |
| **Fraud** | 34 | 114 |
|  | Not Fraud | Fraud |

Actual (rows) / Predicted (columns)

We have used numpy library in python to implement float16 bits data type on the csv file which reduced the size of data for processing; stored in 32-bits to only 16-bits. This is the central part of the thesis through which we are proving that we are achieving machine scalability for credit card fraud detection systems.

## 5.2 Statistical Analysis Results

Below is the demonstration of the results we have obtained after running the code for implementing the float16 bits data type and compare the results to understand if machine learning scalability was attained.

I. **Confusion Matrix for Unbalanced Dataset**

As shown in the below graph, when we test the model on unbalanced data after using float16 data type, the value of confusion matrix has been changed. The negative value has slightly increased more as the dataset was unbalanced and changing the bits to float16 has impacted the confusion matrix.

*Table 12: Confusion Matrix with float 16 on unbalanced data*

## II. Confusion Matrix for Balanced Dataset – SMOTE

To ensure that data set is balanced, we have used the over-sampling technique SMOTE. The ratio of negative is comparatively less.

| | | Predicted | |
|---|---|---|---|
| | | **Not Fraud** | **Fraud** |
| **Actual** | **Not Fraud** | 85275 | 20 |
| | **Fraud** | 27 | 121 |

*Table 13: Confusion Matrix with float16 on balanced data*

## III. Accuracy & Other Parameters

We will calculate the accuracy, precision & recall for both unbalanced and balanced data showing the difference with float16 and without float16 bits.

| Random Forest Classifier (Float16) | |
|---|---|
| **Accuracy** | 0.9995084442259752 |

| Precision | 0.9996015 |
|---|---|
| Recall | 0.99990621 |

| Random Forest Classifier (Without Float16) | |
|---|---|
| Accuracy | 0.999520148 |
| Precision | 0.99962493 |
| Recall | 0.99989448 |

*Table 14 & 14: Both tables are showing the accuracy & other parameters of unbalanced data on float16 bits*

| Random Forest Classifier (SMOTE – Float16) | |
|---|---|
| Accuracy | 0.9994499256814484 |
| Precision | 0.99968348 |
| Recall | 0.99976552 |

| Random Forest Classifier (SMOTE) | |
|---|---|
| Accuracy | 0.9996137776 |
| Precision | 0.99963668 |
| Recall | 0.99997655 |

*Table 15 & 15: Both tables are showing the accuracy & other parameters of balanced data on float16 bits*

IV. **CPU Execution Time – Before Float16 Implementation**

We will show different graphs depicting the difference between CPU execution time on unbalanced and balanced data.

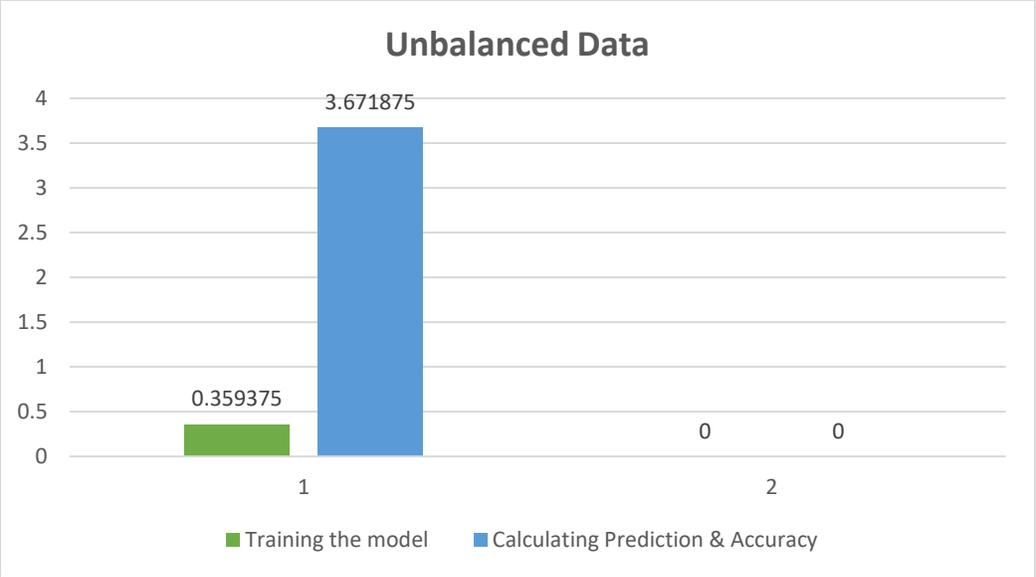

*Table 16: CPU Execution Time graph for training and calculation on unbalanced data*

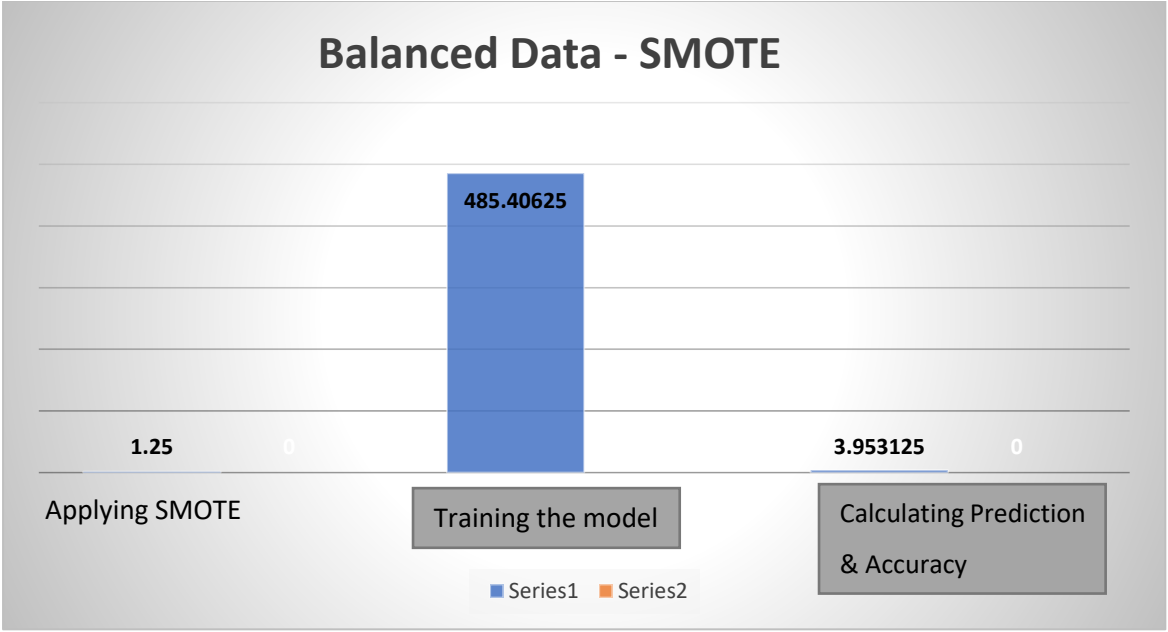

*Table 17: CPU Execution Time on graph for training and calculation on unbalanced data*

V.  **CPU Execution Time – After Float16 Implementation**

We will show different graphs depicting the difference between CPU execution time on unbalanced and balanced data with float16 bits.

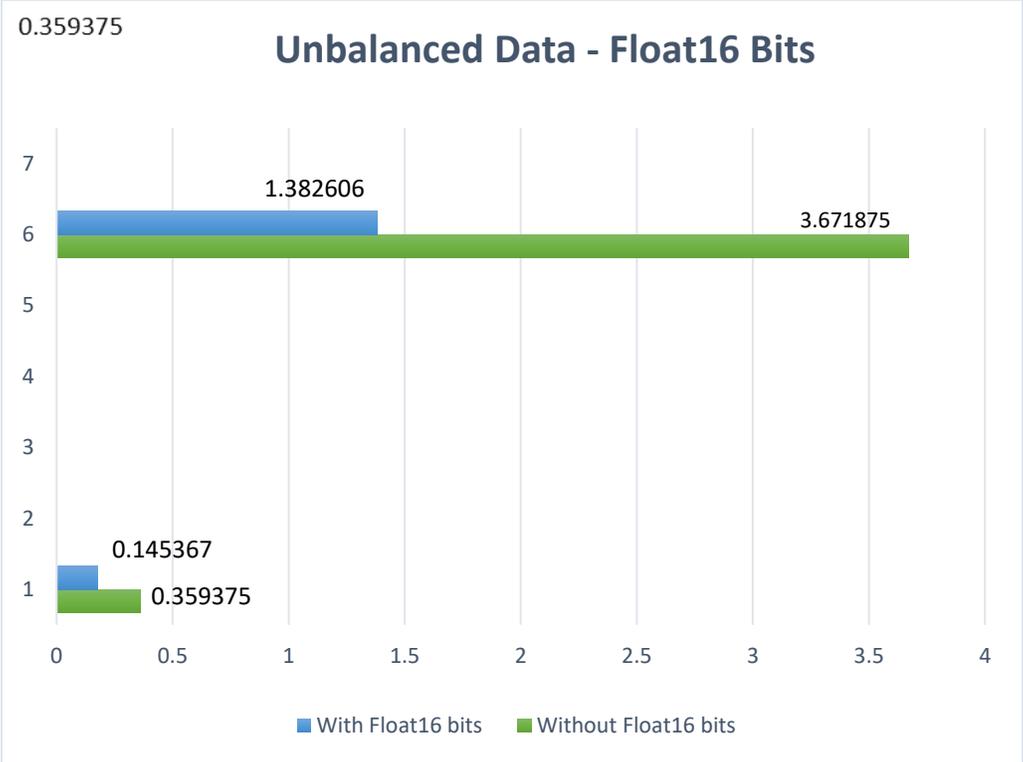

*Table 18: Depiction of CPU execution time with float16 (unbalanced data)*

As shown above, the CPU execution time taken with float16 is almost 55% less than without the float16 bits.

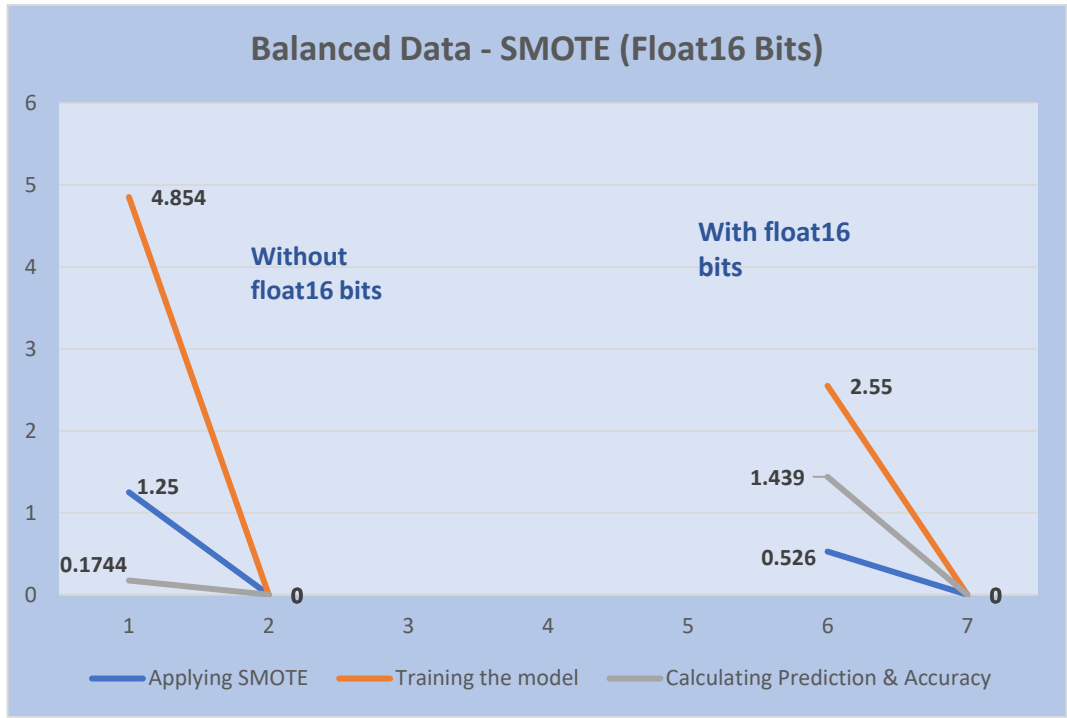

*Table 19: Depiction of CPU execution time with float16 (balanced data)*

As clearly depicted in the above graph, we can deduce that by using float16 bits we can achieve machine learning scalability by reducing the CPU execution time which will result in faster processing and utilization of less resources.

## 6. Conclusion

The thesis aimed to implement a machine learning based scalability for banks credit card fraud detection system. We can conclude based on the results and comparison depicted that execution time was reduced by 50%-60% after we implemented float16 bits on the data. The results also indicate that the accuracy was reduced after implementing float16 bits, but it was only reduced by 0.0002% for unbalanced data and balanced data.

Considering the prompt action required in detecting credit card fraud we can say assume that accuracy can be compromised as it is more important to quickly identify the fraudster before he steals the amount. A fraudster can empty the credit card in matter of seconds before we can detect that a fraud has occurred. This technique can help save millions as it will rapidly detect frauds in credit card.

## 7. Future Work

We can implement two more techniques and resources to further enhance fraud detection of credit cards.

### I. Bfloat16

We can use the **brain floating-point format (bfloat16 or BF16)** which will use 16-bits to represent the floating-point number. The first is used for sign bit, the next 8-bits are used for exponent and the last 7-bits for fraction/mantissa.

The bfloat16 is a better option than selecting the standard fp16 bits because it has more exponents bits than fp16 which has only 5 exponent bits. It also presents accurately all the integers, and they are quite easy to convert and from IEEE-754 standard 32-bits format.

### II. TPUs (Tensor Processing Units)

Since we had a time and resource constraint for the thesis, limited data was used for implementation. If real-time data processing is required for billions of transactions, we can use TPUs provided by GCP (Google Cloud Platform). Currently use of TPUs is not given to every user and it is paid, large banks and organizations can use TPUs to further enhance the data processing time.